\documentclass[10pt,twocolumn,letterpaper]{article}

\usepackage{cvpr}
\usepackage{times}
\usepackage{epsfig}
\usepackage{graphicx}
\usepackage{amsmath}
\usepackage{amssymb}
\usepackage[font=small,skip=2pt]{caption}
\usepackage[font=small]{subfig}
\usepackage{tabularx}
\usepackage{multirow}
\usepackage{soul}

\usepackage[skip=5pt]{caption}

\newcolumntype{Y}{>{\centering\arraybackslash}X}

\usepackage[pagebackref=true,breaklinks=true,letterpaper=true,colorlinks,bookmarks=false]{hyperref}

\cvprfinalcopy 

\def\cvprPaperID{9203} 

\ifcvprfinal\fi
\begin{document}
	
%

\title{Optical Flow in Dense Foggy Scenes using Semi-Supervised Learning}

\author{Wending Yan$^{*1}$, Aashish Sharma$^{*1}$, and Robby T. Tan$^{1,2}$\\
$^1$National University of Singapore, $^2$Yale-NUS College\\
{\tt\small 
	eleyanw@nus.edu.sg, aashish.sharma@u.nus.edu, robby.tan@\{nus,yale-nus\}.edu.sg}
}

\maketitle
\def\thefootnote{*}\footnotetext{Both authors contributed equally to this work.}\def\thefootnote{\arabic{footnote}}
\def\thefootnote{$\dagger$}\footnotetext{This work is supported by   MOE2019-T2-1-130.}\def\thefootnote{\arabic{footnote}}

\begin{abstract}
In dense foggy scenes, existing optical flow methods are erroneous. This is due to the degradation caused by dense fog particles that break the optical flow basic assumptions such as brightness and gradient constancy.
To address the problem, we introduce a semi-supervised deep learning technique that employs real fog images without optical flow ground-truths in the training process. 
Our network integrates the domain transformation and optical flow networks in one framework. 
Initially, given a pair of synthetic fog images, its corresponding clean images and optical flow ground-truths, in one training batch we train our network in a supervised manner. 
Subsequently, given a pair of real fog images and a pair of  clean images that are not corresponding to each other (unpaired), in the next training batch, we train our network in an unsupervised manner.
We then  alternate the training of synthetic and real data iteratively.
We use real data without ground-truths, since to have ground-truths in such conditions is intractable, and also  to avoid the overfitting problem of synthetic data training, where the  knowledge learned on synthetic data cannot be generalized to real data testing.
Together with the network architecture design, we propose a new training strategy that combines supervised synthetic-data training and unsupervised real-data training. 
Experimental results show that our method is effective and outperforms the state-of-the-art methods in estimating optical flow in dense foggy scenes.   
\end{abstract}

\section{Introduction}

Fog is a common and inevitable weather phenomenon. It degrades visibility by weakening the background scene information, and  washing out the colors of the scene. This degradation breaks the Brightness Constancy Constraint (BCC) and Gradient Constancy Constraint (GCC) used in existing optical flow methods. To our knowledge, none of the existing methods can handle dense foggy scenes robustly.
This is because most of them (e.g.~\cite{DeepFlow_2013,FlowNet,FlowNet2,PWC,SpyNet,DCFlow}) are designed under the assumption of clear visibility.

One of the possible solutions is to  render synthetic fog images based on the commonly used physics model (i.e., the Koschmieder model~\cite{koschmieder1924theorie}), and then to train a network on the synthetic fog images and their corresponding optical flow ground-truths in a supervised manner. While in our investigation, it works to some extent, when applied to real dense fog images in the testing stage, it does not perform adequately. The main cause is the domain gap between the synthetic and real fog images. The synthetic images are too crude to represent the complexity of real fog images.
This problem can be fixed by using real fog images, instead of synthetic fog images for training. However, to obtain the correct optical flow ground-truths for real fog images is extremely challenging~\cite{bijelic2018benchmark}.

\begin{figure}
	\begin{center}
        \subfloat[Input Image]{\includegraphics[width=0.235\textwidth]{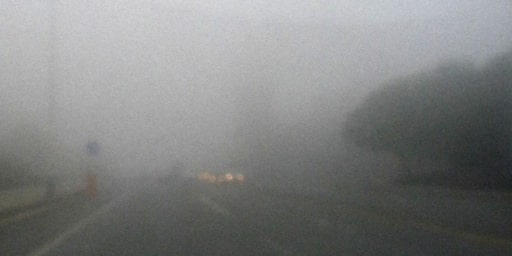}}\hfill
		\subfloat[PWCNet~\cite{PWC}]{\includegraphics[width=0.235\textwidth]{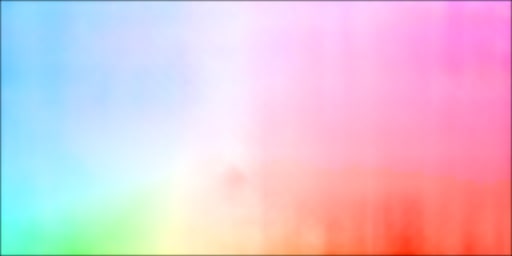}}\hfill
		\vspace{-0.4cm}
		\subfloat[\textbf{Our Result}]{\includegraphics[width=0.235\textwidth]{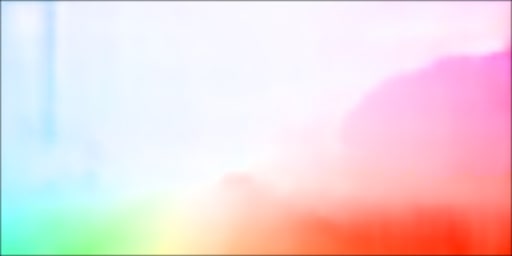}}\hfill
		\subfloat[Ground-Truth]{\includegraphics[width=0.235\textwidth]{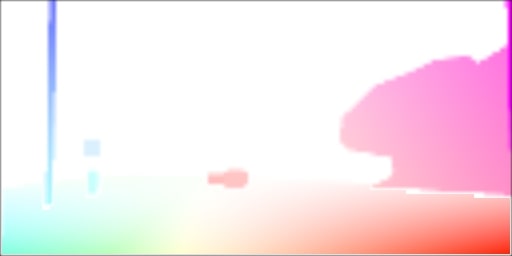}}\hfill
		\vspace{-0.4cm}
	\end{center}
	\caption{(a) Input dense foggy image (first frame). (b) Optical flow result from the existing baseline method PWCNet~\cite{PWC}. We can observe that the result is erroneous and the method cannot handle dense fog. As shown in (c), compared to it, our method performs more robustly.}
	\label{fig:trailer}
	\vspace{-0.15in}
\end{figure}

Another possible solution is to defog the real fog images using an existing defogging method (e.g.,~\cite{tan2008visibility,fattal2008single,he2011single,berman2016non,zhang2018densely,li2018single}), and then to estimate the flow using an existing optical flow method.  This two-stage solution, however, is not effective either. First, existing defogging methods are not designed for optical flow, hence their outputs might not be optimum for flow computation. Second, defogging, particularly for dense fog, is still an open problem. Hence, the outputs of existing defogging methods are still inadequate to make the estimation of optical flow accurate and robust.

Our goal in this paper is to estimate the optical flow from dense fog images robustly. To achieve the goal, we introduce a novel deep learning method that integrates domain transformation (e.g.~\cite{isola2017image,zhu2017unpaired}) and optical flow estimation in a single framework. 
Initially, given a pair of synthetic fog images, its corresponding clean images and optical flow ground-truths, in one training batch, we train our network in a supervised manner. 
Subsequently, given a pair of real fog images and a pair of clean images that are not corresponding to each other (unpaired data), in the next training batch, we train our network in an unsupervised manner. The training of synthetic and real data are carried out alternately. 
We use the synthetic data with ground-truths to guide the network to learn the transformation correctly, so that we can mitigate the fake-content generation problem that is commonly observed in unpaired training~\cite{zhu2017unpaired,UNIT,sharma2019depth}.
We use the real data without ground-truths to avoid the overfitting problem of the synthetic data training, where the knowledge learned from synthetic data in the training cannot be generalized to real data in the testing.

In essence, our method is a semi-supervised method.  
Our domain transformation  enables our network to  learn directly from real data without ground-truths.
When the training input is a pair of clean images, our domain transformation renders the corresponding foggy images, and our optical flow module estimates the flow map. 
Moreover, from the rendered foggy images, our optical flow module also estimates the flow map. Hence, these two different flow maps must be identical. If they are not, then we can backpropagate the error. The same mechanism applies when the training input is a pair of foggy images.
Another advantage of our architecture is that the transformation and optical flow modules can benefit each other: Our domain transformation helps our optical flow module reconstruct the fog-invariant cost volume, and our optical flow module enables our domain transformation module to distinguish some objects from the background through the flow information. 
As a summary, here are our contributions:  
\begin{itemize}
	\setlength{\itemsep}{0pt}
	\setlength{\parskip}{0pt}
	\item We introduce an architecture that integrates  domain transformation and optical flow modules in one framework. The two modules work in mutual cooperation benefiting each other at the feature pyramid levels. 
	\item We propose a training strategy combining synthetic data with ground-truths, clean and fog real data without ground-truths in one integrated learning process.
	\item We provide a domain adaptive method, which can predict optical flow from both clean and fog images. We also show the effectiveness of using photometric and hazeline~\cite{8540862} constraints to make our network learn better about optical flow and fog.
\end{itemize}


\section{Related Work}

Many methods have been proposed to tackle optical flow estimation (\cite{fortun2015optical} for a comprehensive survey). More recently, deep learning is widely used in optical flow methods. Dosovitskiy et. al.~\cite{FlowNet} design FlowNetS and FlowNetC based on the U-Net architecture~\cite{ronneberger2015u}. Their method is a pioneer work in showing the possibility of using a deep-learning method to solve the optical flow problem. Ilg et al.~\cite{FlowNet2} design FlowNet2 by  stacking multiple FlowNetS and FlowNetC networks.  FlowNet2 is trained in a stack-wise manner, and thus is not end-to-end. Sun et al.~\cite{PWC} propose PWCNet. Its performance is comparable to FlowNet2, yet it is  significantly smaller in terms of network parameters. All these methods are fully supervised and trained using synthetic data. In contrast, our method uses semi-supervised learning, employing labeled  synthetic and unlabeled  real data. 

Jason et al.~\cite{jason2016back} propose an unsupervised learning method for flow estimation, for the first time. Ren et al.~\cite{ren2017unsupervised} publish a method with a more complex structure. These two methods simply use the brightness constancy and motion smoothness losses. Some other methods combined depth, ego-motion and optical flow together, such as Yin and Shi~\cite{yin2018geonet} and Ranjan et al.~\cite{ranjan2019competitive}.  These methods, however, use three independent networks to estimate depth, ego-motion and optical flow, and require camera calibration. Generally, the performance of the current unsupervised methods cannot be as accurate and sharp as that of the fully supervised methods. 
To take the advantages of both fully supervised and unsupervised learning, Lai et al.~\cite{lai2017semi} proposed a semi-supervised method, which uses the discriminative loss from the warping difference between two frames.  
Recently, there is progress in unsupervised optical flow (e.g.~\cite{zhu2019unsupervised,wang2019unos,ranjan2019competitive}), under the assumption that the input images are clean. Liu at al. \cite{liu2019selflow} propose a self-supervised method for learning optical flow from unlabeled data. They use photometric loss to obtain reliable flow estimations, which are later used as ground-truths for training.
To our knowledge, none of these methods are designed to handle dense foggy scenes. 
While some previous non-learning-based works (e.g.~\cite{molnar2010illumination}) can handle illumination variations in the images, these methods also cannot handle dense foggy scenes. This is because fog is more than just intensity/illumination changes in the images, and robustness to illumination variations does not necessarily ensure robustness to fog.

Some works address the problem of semantic segmentation under fog (e.g.~\cite{DSHV18,DSHV19}). However, they employ a gradual learning scheme, where the network is first trained on labeled synthetic fog data. Then, the network is used to generate flow results on light real fog data. The network is then trained again on labeled synthetic fog data and light fog real data, for which the results predicted before are used as ground-truths. The entire process is repeated for dense fog real data. While this learning scheme is simple to implement, it has a few problems. First, it makes the entire learning scheme manual. In contrast, our method is completely end-to-end trainable and requires no manual intervention. Second, the results predicted for real data in the previous stage are used as ground-truths for training in the next stage, which could be erroneous. This can lead the network to learning inaccurate flow estimations. In contrast, our method uses accurate flow ground truths from synthetic data, to learn on rendered real fog data. This ensures that the flow network always learns from correct flow ground-truths. 

One possible solution of estimating optical flow in foggy scenes is a two-stage solution: defog first and optical flow estimation afterwards. Many methods in defogging/dehazing have been proposed. (see \cite{li2017haze} for a comprehensive review). A few methods are based on deep learning, e.g.~\cite{cai2016dehazenet,ren2016single,li2017aod,li2018single}. All these methods are based on a single image, and thus can cause inconsistent defogging outputs, which in turn causes the violation of the BCC and GCC. Moreover, defogging,  particularly for dense fog is still an open problem. Hence, the outputs of existing defogging methods can still be inadequate to generate robust optical flow estimation. 


\begin{figure*}[t!]
	\vspace{-0.3cm}
	\begin{center}
		\captionsetup[subfigure]{labelformat=empty}
		\subfloat{\includegraphics[width=1.0\textwidth]{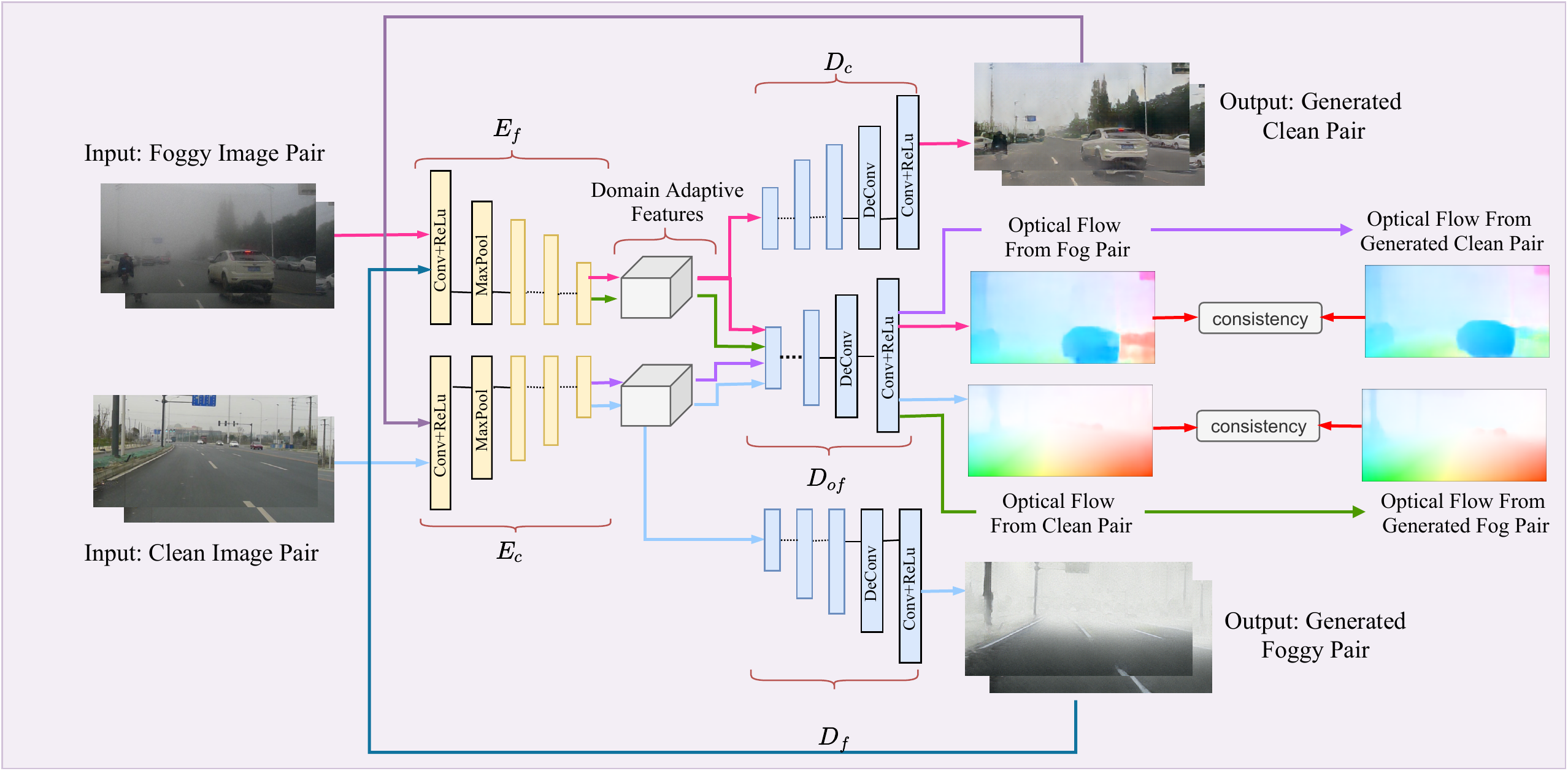}}\hfill	
		\vspace{-0.4cm}
	\end{center}
	\caption{Overall architecture of our network.}
	\label{fig:network}
	\vspace{-0.3cm}
\end{figure*}

\section{Proposed Method}

\subsection{Network Architecture} 

\paragraph{Optical Flow Network}
Our optical flow module consists of two encoders $E_f$, $E_c$ and a decoder $D_{of}$, which are shown in Fig.~\ref{fig:network}, where subscripts $f$, $c$, and $of$ stand for fog, clean, and optical flow, respectively. The two encoders ($E_f$ and $E_c$,) extract features from the fog and clean input images respectively. 
They have the same architecture, but independent weights. As recent works~\cite{SpyNet,PWC} show that pyramid features improve the estimation of optical flow, we design our encoders in the same way. Our decoder correlates the pyramid features from two input images to form a cost volume, which is used to predict optical flow. 
Since our decoder receives features from the two encoders working on different domains (fog and clean), it encourages the two encoders to generate domain adaptive features. 
This domain adaptation ensures that robust optical flow is generated from the two domain inputs. 

\vspace{0.3cm}
\noindent {\bf Domain Transformation Network} 
Our domain transformation module is formed by the encoders, $E_f$ and $E_c$,  and two decoders, $D_f$ and $D_c$. The fog encoder, $E_f$, takes the fog images as the input, and outputs feature pyramids.  The clean decoder, $D_c$, processes the features, and constructs the clean version of the input images. The other encoder, $E_c$, does the same, however instead of fog images, it takes clean images as the input. The fog decoder, $D_f$, processes the features produced by  $E_c$, and transforms them to fog images. To ensure the proper quality of our transformed clean and fog images, we employ the discriminative loss \cite{goodfellow2014generative}.
While domain transformation is not our main goal, the quality of the transformed images can affect the optical flow result.
Note that,  we employ feature pyramids in computing the features, so that the same features can also be used by our optical flow network.

\subsection{Semi-Supervised Training Strategy}
\label{sec:ssl}
To train our network, ideally we should use real fog data with the corresponding optical flow ground-truths. Unfortunately, to obtain the ground-truths of real fog images is extremely intractable. The best possible technology we can employ currently is LIDAR sensors. However, LIDAR captures only sparse depths and stationary objects. Moreover, it has limited depth range and its accuracy is affected by fog dense particles~\cite{bijelic2018benchmark}. 
An alternative solution is to use synthetic fog images, whose corresponding optical flow is easy to obtain. However, it is known that there are significant gaps between synthetic and real fog images. Synthetic fog images are too simplistic and cannot represent real fog and its complexity in many conditions. Because of these problems, we utilize real clean (no fog) images to help our network learn about fog, clean background scenes, and optical flow. While there are domain gaps between clean real images and fog real images, we bridge the gaps through our domain transformation network.

Our training strategy includes datasets both with and without ground-truths, involving real fog images, synthetic fog images, and real clean images. The reason we use the synthetic fog images is because, they can help guide the network to transform the features from different image domains more correctly by mitigating the generation of fake contents during the transformation.
The whole process of our training strategy can be separated into three stages: Synthetic-fog training stage, real-clean training stage, and real-fog training stage.

\subsection{Synthetic-Data Training Stage}
\label{sec:synthetic_training}
Given synthetic fog images, their corresponding synthetic clean background images, and their corresponding optical flow ground-truths, we can train our network in a fully supervised manner. First, to train the optical flow module: $\{E_f,E_c, D_{of}\}$, we use EPE (End-Point Error) losses between the predicted optical flow and the corresponding ground-truths for both synthetic fog and clean input images:
\begin{eqnarray}
\mathcal{L}_{\text{EPE}^f_s}(E_f,D_{of}) &=& 
\mathbb{E}_{(x^f_{s1}, x^f_{s2})}\big[\lVert \widehat{of}^f - of^f_{gt}\rVert_2\big], \label{eq:EPE_f_s} \\ 
\mathcal{L}_{\text{EPE}^c_s}(E_c,D_{of}) &=& 
\mathbb{E}_{(x^c_{s1}, x^c_{s2})}\big[\lVert \widehat{of}^c - of^c_{gt}\rVert_2\big],\label{eq:EPE_c_s} 
\end{eqnarray}
with: 
\begin{eqnarray}
\widehat{of}^f &=& D_{of}[(E_f[x^f_{s1}],E_f[x^f_{s2}])],\\ 
\widehat{of}^c &=& D_{of}[(E_c[x^c_{s1}], E_c[x^c_{s2}])],
\end{eqnarray}
where $(x^f_{s1}, x^f_{s2})$ and $(x^c_{s1}, x^c_{s2})$ are the synthetic fog and synthetic clean image pairs. $of^f_{gt}$ and $of^c_{gt}$ are the optical flow ground-truths of the synthetic fog and synthetic clean images respectively. 

To train the domain transformation module:  $\{E_f,D_c\}$  and $\{E_c,D_f\}$, we define L1 losses:
\begin{eqnarray}
\mathcal{L}_{L1^f_s} (E_f,D_c) &=& \mathbb{E}_{x^f_{s}}\big[\lVert  \hat{x}^c_s - x^c_{gt}\rVert_1\big],\label{eqn_L1_f_s} \\ 
\mathcal{L}_{L1^c_s}(E_c,D_f) &=& \mathbb{E}_{x^c_{s}}\big[\lVert \hat{x}^f_s - 
x^f_{gt}\rVert_1\big],
\label{eqn_L1_c_s}
\end{eqnarray}
where, $\hat{x}^c_s =  D_{c}[(E_f[x^f_{s}]])]$, 
and 
$\hat{x}^f_s = D_{f}[(E_c[x^c_{s}]])]$
are the rendered clean and fog images, respectively.
$x^c_{gt}$, $x^f_{gt}$ are the synthetic clean and synthetic fog ground-truth images, respectively. 
In addition, we also apply the discriminative loss~\cite{goodfellow2014generative} to ensure that the transformations from clean-to-fog images and from fog-to-clean images are consistent with the appearance of synthetic fog and synthetic clean images.

\subsection{Real Clean Data Training Stage}
\label{sec:clean_training}
In this stage, we use the real clean images without optical flow ground-truths and without real fog image ground-truths to train the network. As shown in the second row of Fig.~\ref{fig:network}, first, we compute the optical flow directly from the input real clean images, $x^c_{r1},x^c_{r2}$:
\begin{eqnarray}
\widehat{of}^c =  D_{of}[(E_c[x^c_{r1}], E_c[x^c_{r2}]).
\end{eqnarray}
Concurrently, we transform the input clean images, $x^c_{r1}, x^c_{r2}$ to fog images, $\hat{x}^f_{r1},\hat{x}^f_{r2}$:
\begin{eqnarray}
\hat{x}^f_{r1}&=&D_f[E_c[x^c_{r1}]], \\ 
\hat{x}^f_{r2}&=&D_f[E_c[x^c_{r2}]].
\end{eqnarray}
From the rendered fog images, $\hat{x}^f_{r1},\hat{x}^f_{r2}$, subsequently we transform them further to obtain the rendered clean images, $\hat{\hat{x}}^c_{r1},\hat{\hat{x}}^c_{r2}$:
\begin{eqnarray}
\hat{\hat{x}}^c_{r1}&=&D_c[E_f[\hat{x}^f_{r1}]],\\
\hat{\hat{x}}^c_{r2}&=&D_c[E_f[\hat{x}^f_{r2}]].
\end{eqnarray}
At the same time, we also compute the optical flow from the rendered fog images, $\hat{x}^f_{r1},\hat{x}^f_{r2}$:
\begin{eqnarray}
\widehat{\widehat{of^f}} &=& D_{of}[\hat{x}^f_{r1},\hat{x}^f_{r2}]. 
\end{eqnarray}

The whole process above, from the input real clean images, $x^c_{r1},x^c_{r2}$ to the rendered clean, 
$\hat{\hat{x}}^c_{r1},\hat{\hat{x}}^c_{r2}$, 
and to the estimated optical flow, $\widehat{of}^c$ and  $\widehat{\widehat{of^f}}$ is a feedforward process. 
Initially, we rely on the network's weights  learned from synthetic data for this feedforward process. To refine the weights, we train the network further using our current real data. The training is based on a few losses: Transformation consistency, EPE, discriminative, and hazeline losses.

\vspace{0.3cm}
\noindent{\bf Transformation Consistency Loss}
To train the domain transformation modules: $E_f, E_c, D_f, D_c$, 
we define our consistency loss between the clean input images, $x^c_{r1}, x^c_{r2}$, and the rendered clean images, $\hat{\hat{x}}^c_{r1}, \hat{\hat{x}}^c_{r2}$, as:
\begin{multline}
\mathcal{L}_{\text{CON}^c_r}(E_f,E_c,D_f,D_c) \\=
\mathbb{E}_{x^c_r}
\big[\lVert x^c_{r1} - \hat{\hat{x}}^c_{r1} \rVert_1 
+ 
\lVert x^c_{r2} - \hat{\hat{x}}^c_{r2} \rVert_1\big].
\end{multline}               
This loss is a pixel-wise computation, since the real clean and rendered clean images must share the same optical flow up to the pixel level. In this  backpropagation process, we keep $D_{of}$ frozen.

\vspace{0.3cm}
\noindent{\bf EPE Loss}
Since we do not have the optical flow ground-truths of the real clean input images, to train our modules $E_f$ and $D_{of}$, we define the EPE loss by comparing the predicted optical flow from the real clean input images and the predicted optical flow from the rendered fog images:
\begin{eqnarray}
&\mathcal{L}_{\text{EPE}^c_r}(E_f,D_{of}) = \mathbb{E}_{(x^c_{r1}, x^c_{r2})}\big[\lVert 
\widehat{of}^c, \widehat{\widehat{of^f}}
\lVert_2\big],
\end{eqnarray}
where $\widehat{of}^c, \widehat{\widehat{of^f}}$ are the predicted optical flow fields from the input clean images, and from the rendered fog images, respectively.
During the backpropagation of this EPE loss, only $E_f$ and $D_{of}$ are updated, and the rest remain frozen.

\vspace{0.3cm}
\noindent{\bf Discriminative Loss}
 To train the transformation modules, $E_c,D_f$, we use the discriminative loss~\cite{goodfellow2014generative} to ensure that the rendered fog images look as real as possible (since we do not have the corresponding real-fog ground-truths). 
 For this purpose, we define our discriminative loss as: 
\begin{align}
\vspace{-0.15in}
\mathcal{L}_{\text{GAN}^c_r}(E_c,D_f) =\mathbb{E}_{x^c_r}\big[(\log(1- Dis_f[D_{f}[(E_c[x^c_{r}]]])\big],
\label{eqn_GAN_c_r}
\end{align}            
where $Dis[.]$ is our discriminative module, which assesses the outputs of $D_f$. We keep other modules frozen, while updating the weights of $E_c,D_f$.

\vspace{0.3cm}
\noindent{\bf Hazeline Loss}
Since we do not have the ground-truths of the corresponding real fog images, applying the discriminative loss alone will be insufficient to train the modules $E_c,D_f$ properly. Improper training can cause the generation of fake contents~\cite{zhu2017unpaired,UNIT,sharma2019depth}. The guidance of the synthetic training data (Sec.~\ref{sec:synthetic_training}) can mitigate the problem; since synthetic fog images are rendered using a physics model, and thus $E_c,D_f$ learn the underlying physics model from the synthetic fog images. To strengthen the transformation even further, we add a loss based on the following physics model~\cite{he2011single,tan2008visibility,berman2016non} (also used in the rendering of our synthetic fog images):
\begin{eqnarray}
x^f(\mathbf{x}) = x^c(\mathbf{x}) \alpha (\mathbf{x})  + (1-\alpha(\mathbf{x})) \mathbf{A},
\label{eq:fog_model}
\end{eqnarray}
where $x^f$ is the fog image, $x^c$ is the clean (no fog) image ground-truth.  $\mathbf{A}$ is the atmospheric light. $\alpha$ is the attenuation factor, and $\mathbf{x}$ is the pixel location.   

Berman et al.~\cite{berman2016non} observe that  in the RGB space, $x^f$, $x^c$, and $\mathbf{A}$ are colinear, due to the linear combination described in the model (Eq.~(\ref{eq:fog_model})). 
Unlike Berman et al.'s method, instead of using the RGB space, we use the 2D chromaticity space \cite{tan2008visibility}; since, there is no robust way to estimate the intensity of the atmospheric light \cite{sulami2014automatic}.
The chromaticity of the clean input image is defined as:
\begin{eqnarray}
\gamma^c_{r,ch} = \frac{x^c_{r,ch}}{x^c_{r,R}+x^c_{r,G}+x^c_{r,B}},
\end{eqnarray}    
where the index $ch = \{R,G,B\}$ is the RGB color channel. Accordingly, the chromaticity of the rendered fog image by $E_c, D_f$ is defined as:
\begin{eqnarray}
\sigma^c_{r,ch} = \frac{\hat{x}^f_{r, ch}}{\hat{x}^f_{r, R}+\hat{x}^f_{r,G}+\hat{x}^f_{r, B}}.
\end{eqnarray}  
Lastly, the atmospheric light chromaticity of the rendered fog image is defined as:
\begin{eqnarray}
\alpha^c_{r,ch} = \frac{A[\hat{x}^f_{r, ch}]}{A[\hat{x}^f_{r, R}]+ A[\hat{x}^f_{r, G}] + A[\hat{x}^f_{r, B}]}, 
\end{eqnarray}
where $A[.]$ is the function that obtains the  chromaticity or color of the atmospheric light. This function is basically a color constancy function, hence any color constancy algorithm can be used~\cite{gijsenij2011computational}. In our implementation, to obtain the atmospheric light chromaticity from fog images, we simply use the brightest patch assumption \cite{tominaga2001scene}. 

Therefore, we define our hazeline loss, which is based on the collinearity in the chromaticity space as:
\begin{align}
\vspace{-0.15in}
\mathcal{L}_{\text{HL}^c_r}(E_c,D_f) =\mathbb{E}_{x^c_r}\big[1 - \frac{(\sigma^c_r - \alpha^c_r) \cdot (\gamma^c_r - \alpha^c_r)}{\lVert(\sigma^c_r - \alpha^c_r)\rVert\lVert(\gamma^c_r - \alpha^c_r)\rVert}\big].\label{eqn_HL_c_r} 
\end{align}           
Like the discriminative loss, while updating the weights of $E_c,D_f$, we keep other modules frozen.

\subsection{Real Fog Data Training Stage}
\label{sec:fog_training}
In this stage, we use the real fog images without optical flow ground-truths and without clean-image ground-truths to train the network. As shown in Fig.~\ref{fig:network}, module $E_f$ takes the fog images, $x^f_{r1},x^f_{r2}$, as the input and generate features, which are used by $D_{of}$ to predict the optical flow:
\begin{eqnarray}
\widehat{of}^f =  D_{of}[(E_f[x^f_{r1}], E_f[x^f_{r2}]).
\end{eqnarray}
$D_{of}$ can handle fog images, since it was trained in the previous stage (Sec.~\ref{sec:clean_training}) using the rendered fog images. At the same time, we transform the input fog images, $x^f_{r1}, x^f_{r2}$ to clean images, $\hat{x}^c_{r1},\hat{x}^c_{r1}$, respectively:
\begin{eqnarray}
\hat{x}^c_{r1}&=&D_c[E_f[x^f_{r1}]], \\
\hat{x}^c_{r2}&=&D_c[E_f[x^f_{r2}]].
\end{eqnarray}
The transformation modules $D_c, E_f$ had been initially trained in the previous stage as well.
From the rendered clean images, $\hat{x}^c_{r1},\hat{x}^c_{r2}$,  we transform them further to obtain the rendered fog images, $\hat{\hat{x}}^f_{r1},\hat{\hat{x}}^f_{r2}$, respectively:
\begin{eqnarray}
\hat{\hat{x}}^f_{r1}&=&D_f[E_c[\hat{x}^c_{r1}]],\\
\hat{\hat{x}}^f_{r2}&=&D_f[E_c[\hat{x}^c_{r2}]].
\end{eqnarray}
We also compute the optical flow from the rendered clean images, $\hat{x}^c_{r1},\hat{x}^c_{r2}$:
\begin{eqnarray}
\widehat{\widehat{of^c}} &=& D_{of}[\hat{x}^c_{r1},\hat{x}^c_{r2}].
\end{eqnarray}
Like in the previous stage, to train the network, we use all the losses we defined in Sec.~\ref{sec:clean_training}, except for the EPE loss.
In this training stage, we still compare the EPE loss between $\widehat{of}^f$ and $\widehat{\widehat{of^c}}$, which we call the optical flow consistency loss~\cite{sharma2019depth}. However, the goal is no longer for estimating flow accurately, but for the two encoders to extract proper domain adaptive features. 
Thus, during the backpropagation of this loss, only two encoders $E_c$ and $E_f$ are updated, and the rest are kept frozen. 
%


\subsection{Photometric Consistency Map}
\label{sec:photo_consistency}
In the second training stage, we use the rendered clean images, rendered fog images, and estimated optical flow together to train our network. 
However, the estimated optical flow might still be inaccurate, which  can affect the learning process of the whole network. 
To address this problem, we generate a binary mask based on the photometric consistency of the estimated optical flows. 
The consistency is computed from the two input clean images and their estimated optical flow.
The consistency is then binarized into a mask, and then we apply the mask to the EPE loss. This enables us to filter out the inaccurate estimations of optical flow during the backpropagation.

\section{Implementation}
\label{sec:implementation}
Our network in total has two encoders, three decoders and two discriminators. Each of the two encoders contains 6 convolution layers. From an input image, each encoder extracts pyramid features at 6 different levels. As a result, the optical flow decoder has a pyramid structure. Its inputs are the five pairs of pyramidal features from a pair of input images. These features are the five deep-layers features extracted by the encoder. The features of each layer are warped based on the previous level of optical flow, and then we compute the cost volume, which is used to estimate optical flow. 
As for the two decoder for the  domain transformation,  the input images, first layer features and second layer features are convoluted into the same shape as the third layer features by a convolution layer. Next, these four features with the same shape are concatenated together, and put into ResNet~\cite{he2016deep}. This ResNet contains six blocks, and its output has the same shape as input. Finally, the deconvolution layers process on the output from ResNet to generate the domain transformation result. The network architecture of each discriminator is similar to that of the PatchGAN discriminator~\cite{isola2017image}, containing five convolution layers.

For training images, we use randomly cropped images of 256x512 resolution. We set the batch size to 3. We use Adam~\cite{adam} for the optimizers of all the modules, and its parameters, $\beta_1$ and $\beta_2$, are set 0.5 and 0.999 respectively. The learning rate is set to 0.0002. All the modules are trained from scratch.
We collected real clean and real fog images. All contain urban scenes. We use the VKITTI dataset~\cite{Gaidon:Virtual:CVPR2016} for rendering synthetic fog images for the fully supervised training, as it has both depth maps and optical flow ground-truths. We specifically select the overcast images (with no sunlight) so that the rendered fog images look more realistic. With the available depth maps, we can generate fog images from VKITTI, with random atmospheric light and attenuation coefficient.  The fog in synthetic data is generated by following the physics model~\cite{koschmieder1924theorie} for fog, expressed in Eq.~(\ref{eq:fog_model}).

\section{Experimental Result}
For evaluation, we compare our method with the following methods:  original FlowNet2~\cite{FlowNet2}, original  PWCNet~\cite{PWC}, which are the two state-of-the-art fully supervised methods; and optical flow network in competitive collaboration (CC)~\cite{ranjan2019competitive} and SelfFlow~\cite{liu2019selflow}, which are two state-of-the-art unsupervised methods; FlowNet2-fog and PWCNet-fog, where we retrain the original FlowNet2 and PWCNet using our synthetic fog images and their optical flow ground-truths; FlowNet2-defog, PWCNet-defog and CC-defog which are two-stage solutions where we combine a defogging method with the original FlowNet2, PWCNet and CC. The defogging method is Berman et al.'s~\cite{8540862}, which is one of the state-of-the-art defogging methods.

\begin{figure*}
	\begin{center}
		\captionsetup[subfigure]{labelformat=empty}
		{\includegraphics[width=0.1428\textwidth]{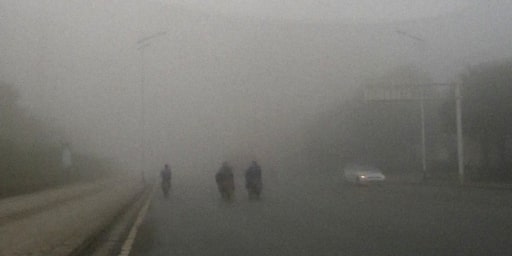}}\hfill
		{\includegraphics[width=0.1428\textwidth]{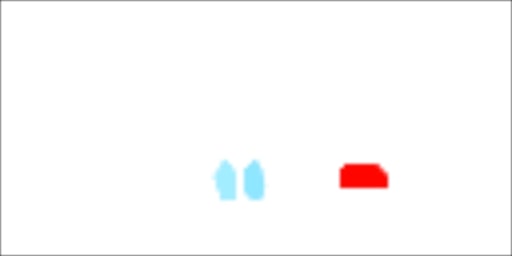}}\hfill
		{\includegraphics[width=0.1428\textwidth]{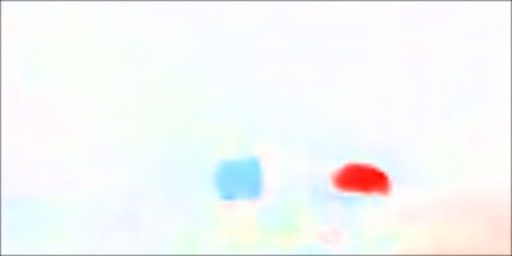}}\hfill
		{\includegraphics[width=0.1428\textwidth]{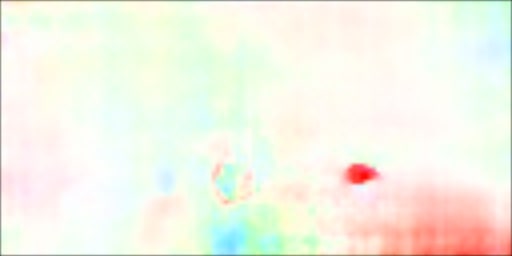}}\hfill
		{\includegraphics[width=0.1428\textwidth]{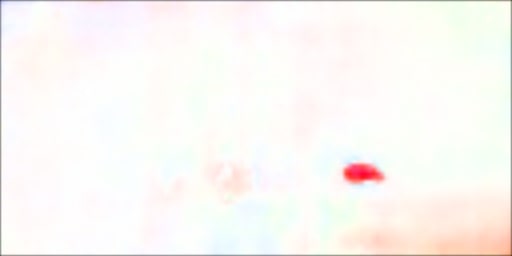}}\hfill 
		{\includegraphics[width=0.1428\textwidth]{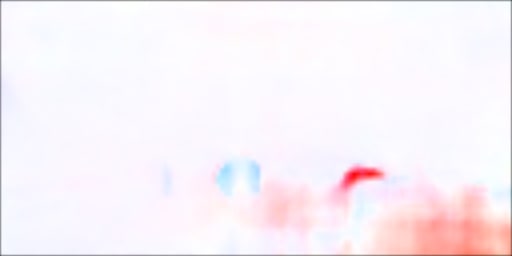}}\hfill 	
		{\includegraphics[width=0.1428\textwidth]{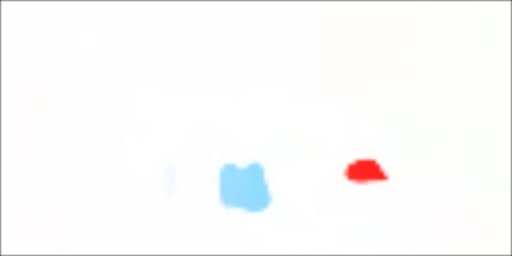}}\hfill		
		
		{\includegraphics[width=0.1428\textwidth]{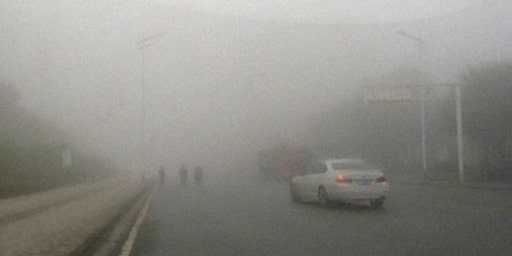}}\hfill
		{\includegraphics[width=0.1428\textwidth]{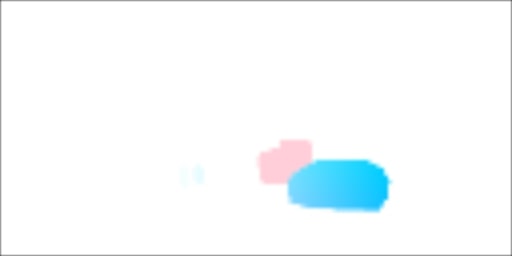}}\hfill
		{\includegraphics[width=0.1428\textwidth]{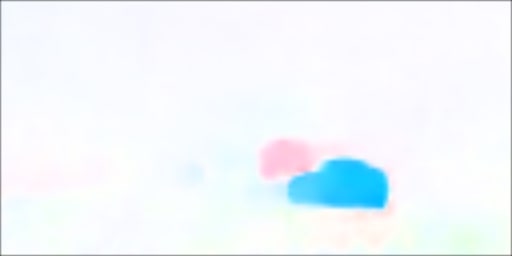}}\hfill
		{\includegraphics[width=0.1428\textwidth]{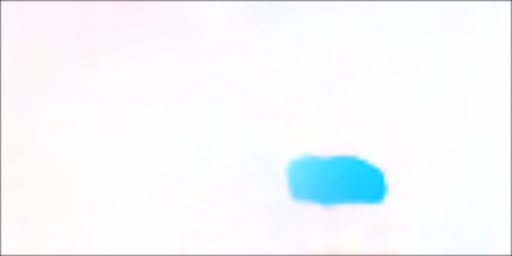}}\hfill
		{\includegraphics[width=0.1428\textwidth]{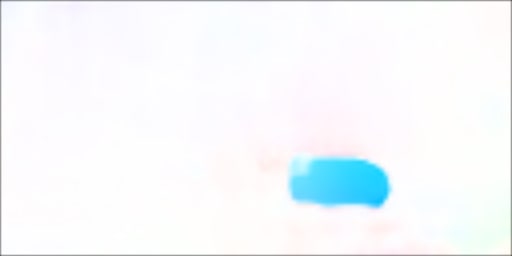}}\hfill 
		{\includegraphics[width=0.1428\textwidth]{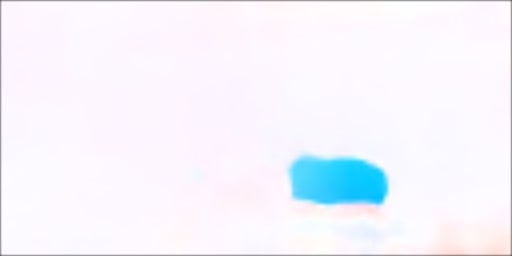}}\hfill	
		{\includegraphics[width=0.1428\textwidth]{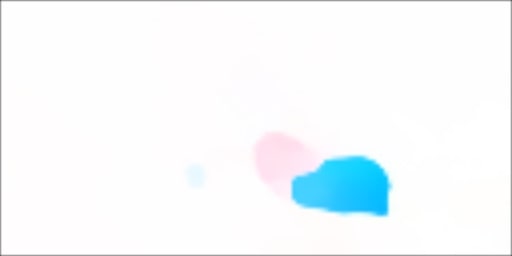}}\hfill	
		
		{\includegraphics[width=0.1428\textwidth]{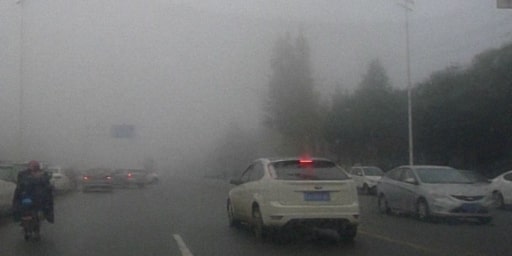}}\hfill
		{\includegraphics[width=0.1428\textwidth]{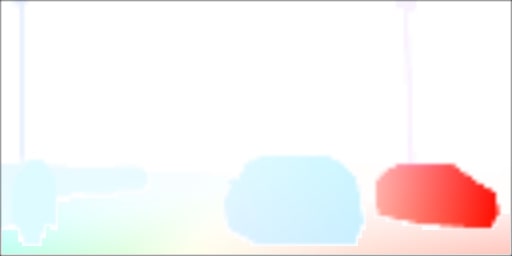}}\hfill
		{\includegraphics[width=0.1428\textwidth]{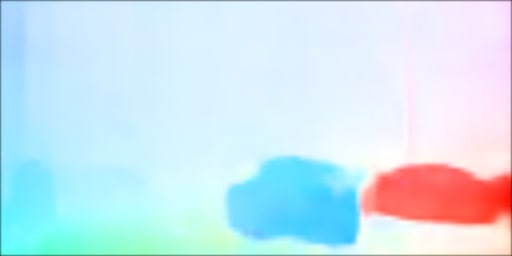}}\hfill
		{\includegraphics[width=0.1428\textwidth]{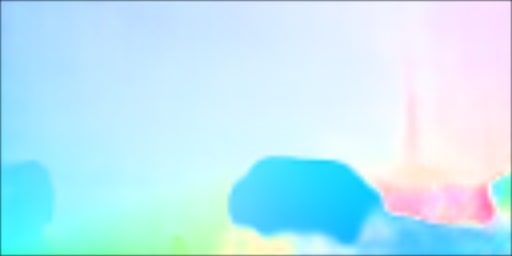}}\hfill
		{\includegraphics[width=0.1428\textwidth]{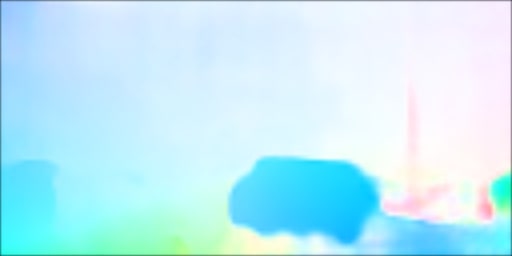}}\hfill 
		{\includegraphics[width=0.1428\textwidth]{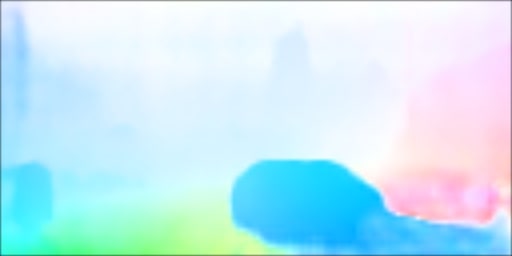}}\hfill
		{\includegraphics[width=0.1428\textwidth]{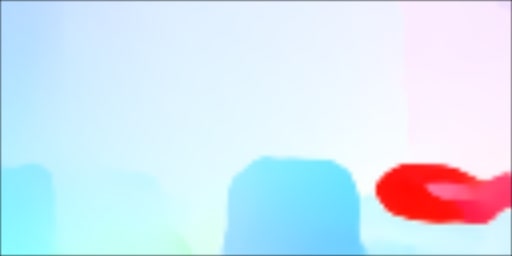}}\hfill	 	
		
		{\includegraphics[width=0.1428\textwidth]{figures/Result_chengdu_foggy/frame1/0079.jpg}}\hfill
		{\includegraphics[width=0.1428\textwidth]{figures/Result_chengdu_foggy/gt/0079.jpg}}\hfill
		{\includegraphics[width=0.1428\textwidth]{figures/Result_chengdu_foggy/ours/0079.jpg}}\hfill
		{\includegraphics[width=0.1428\textwidth]{figures/Result_chengdu_foggy/pwc_orig/0079.jpg}}\hfill
		{\includegraphics[width=0.1428\textwidth]{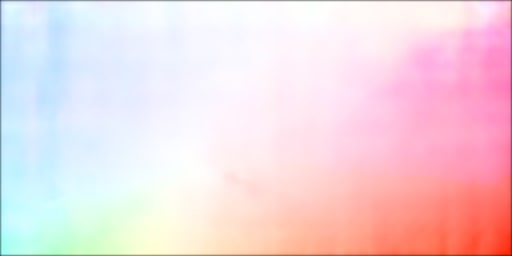}}\hfill 
		{\includegraphics[width=0.1428\textwidth]{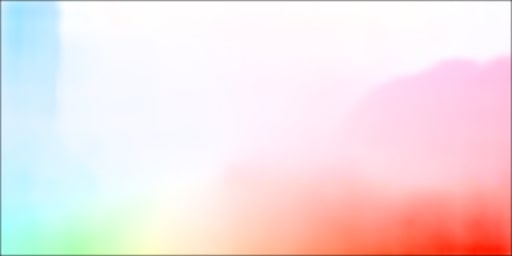}}\hfill
		{\includegraphics[width=0.1428\textwidth]{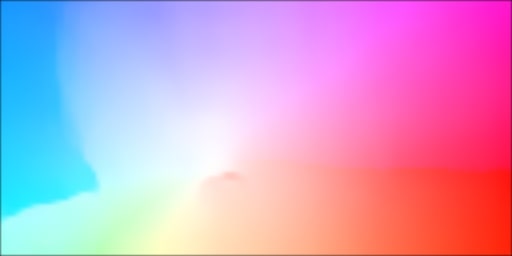}}\hfill
		
		{\includegraphics[width=0.1428\textwidth]{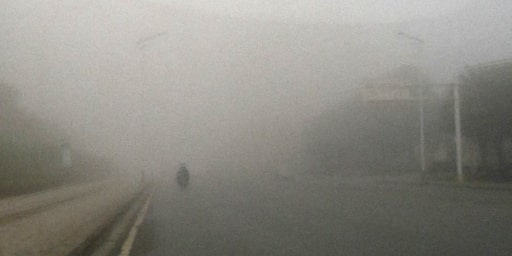}}\hfill
		{\includegraphics[width=0.1428\textwidth]{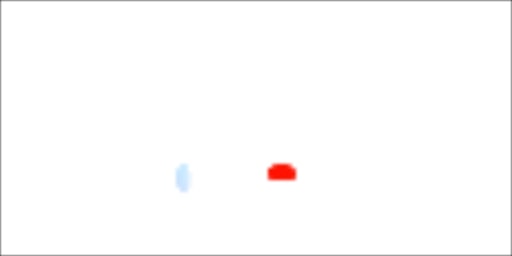}}\hfill
		{\includegraphics[width=0.1428\textwidth]{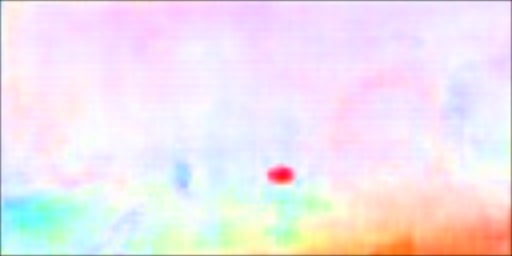}}\hfill
		{\includegraphics[width=0.1428\textwidth]{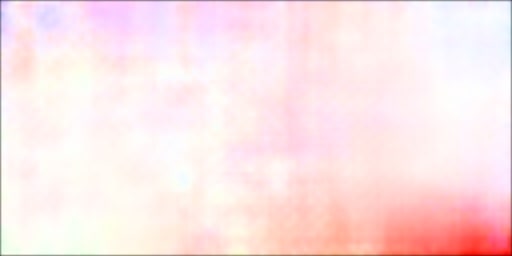}}\hfill
		{\includegraphics[width=0.1428\textwidth]{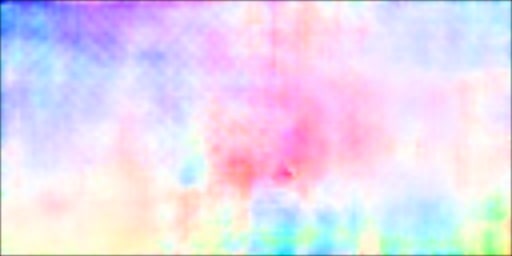}}\hfill 
		{\includegraphics[width=0.1428\textwidth]{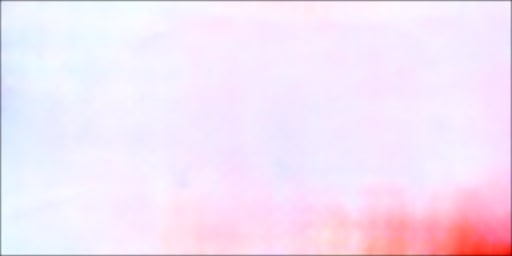}}\hfill
		{\includegraphics[width=0.1428\textwidth]{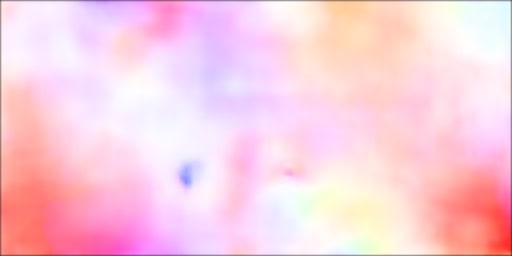}}\hfill
		
		\vspace{-0.14in}
		\subfloat[Input Image]{\includegraphics[width=0.1428\textwidth]{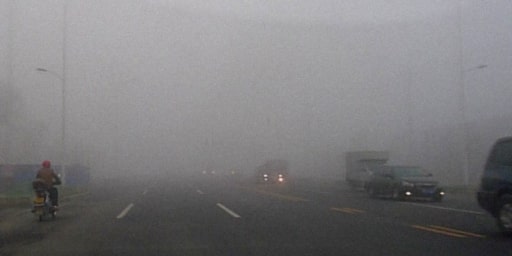}}\hfill
		\subfloat[Ground Truth]{\includegraphics[width=0.1428\textwidth]{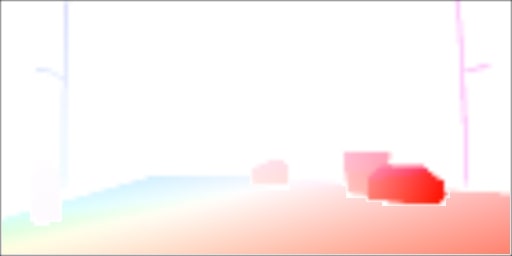}}\hfill
		\subfloat[\textbf{Our Result}]{\includegraphics[width=0.1428\textwidth]{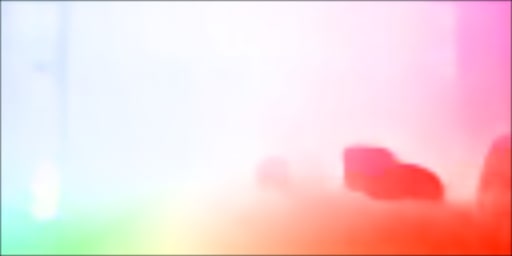}}\hfill
		\subfloat[PWCNet]{\includegraphics[width=0.1428\textwidth]{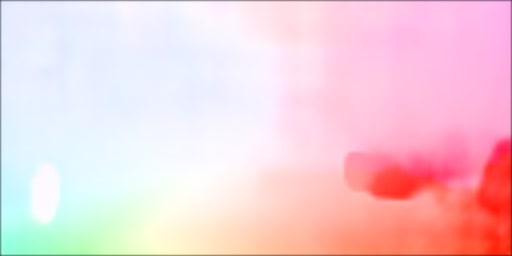}}\hfill
		\subfloat[PWCNet-defog]{\includegraphics[width=0.1428\textwidth]{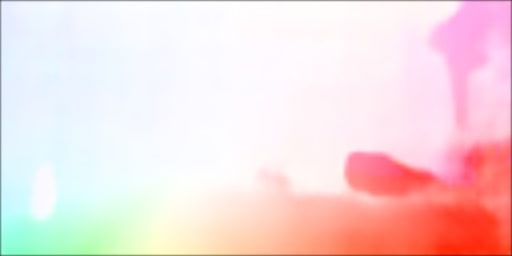}}\hfill 
		\subfloat[PWCNet-fog]{\includegraphics[width=0.1428\textwidth]{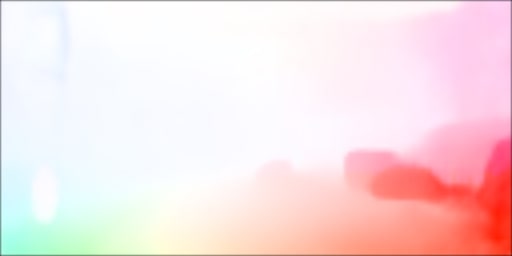}}\hfill
		\subfloat[FlowNet2]{\includegraphics[width=0.1428\textwidth]{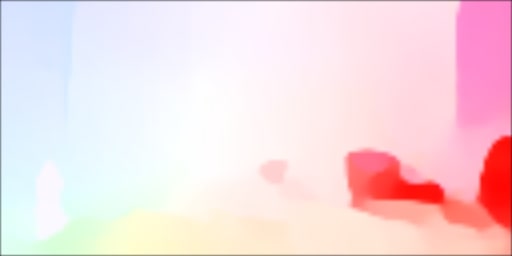}}\hfill
		\vspace{-0.4cm}		
	\end{center}
	\caption{Qualitative comparison of our methods with the state-of-the-art methods and their variants on real fog images.}\label{qual}
\end{figure*}

We use 2,224 real clean and 2,346 real fog image pairs for training. For evaluation, following~\cite{li2019rainflow}, we manually annotate 100 real fog image pairs. The annotated optical flow ground-truths are done for some rigid objects using the manual flow annotation method introduced in~\cite{liu2008human}. We further use 1,770 and 200
 randomly sampled image pairs from the vKITTI dataset for training and validation, respectively. We use other 100 image pairs for testing. We render fog in all the images from the vKITTI dataset using the physics model~\cite{koschmieder1924theorie}, with random atmospheric light and attenuation coefficient (or fog density).

\begin{table}[t!]
	\small
	\centering
	\renewcommand{\arraystretch}{1.25}
	\caption {Quantitative results on our real dense foggy dataset.} \label{table_chengdu_foggy}
	\begin{tabularx}{\columnwidth}{ c|Y|Y|Y }
		\hline
		\hline
		\multirow{2}{*}{Method}  & \multirow{2}{*}{EPE} & \multicolumn{2}{c}{Bad Pixel} \\
		\cline{3-4}
		& & $\delta=3$ & $\delta=5$ \\
		\hline
		CC~\cite{ranjan2019competitive}   &7.56 &76.61\% &45.79\%\\	
		\hline
		CC-defog~\cite{ranjan2019competitive} &11.67 &72.22\% &42.31\%\\			
		\hline	                    
		PWCNet~\cite{PWC} &6.36 &54.90\% &38.47\%\\
		\hline
		PWCNet-defog~\cite{PWC} &6.16 &53.98\% &38.50\%\\
		\hline
		PWCNet-fog~\cite{PWC} &6.10 &56.39\% &39.42\%\\
		\hline                    
		FlowNet2~\cite{FlowNet2} &4.74 &42.06\% &26.75\%\\
		\hline
		FlowNet2-defog~\cite{FlowNet2} &4.72 &43.12\% &26.60\%\\
		\hline
		FlowNet2-fog~\cite{FlowNet2} &5.19 &49.66\% &31.89\%\\
		\hline
		SelfFlow~\cite{liu2019selflow} &6.53 &70.92\% &56.01\% \\
		\hline
		\textbf{Ours} &\textbf{4.32} &\textbf{41.26\%} &\textbf{25.24\%}\\
		\hline
		Ours (no hazeline) &4.82 &43.41\% &31.60\%\\
		\hline
		\hline		
	\end{tabularx}  
\end{table}

\begin{table}[t!]
	\small
	\centering
	\renewcommand{\arraystretch}{1.25}
	\caption {Quantitative results on synthetic foggy vKITTI dataset.} \label{table_vkitti_foggy}
	\begin{tabularx}{\columnwidth}{ c|Y|Y|Y }
		\hline
		\hline
		\multirow{2}{*}{Method}  & \multirow{2}{*}{EPE} & \multicolumn{2}{c}{Bad Pixel} \\
		\cline{3-4}
		& & $\delta=1$ & $\delta=3$ \\
		\hline    
		CC~\cite{ranjan2019competitive}     &7.53 &70.54\% &51.46\%\\	
		\hline
		CC-defog~\cite{ranjan2019competitive} &7.91 &65.51\% &38.90\%\\         
		\hline   
		PWCNet &3.23 & 52.84\% &19.01\% \\
		\hline
		PWCNet-defog~\cite{PWC} &3.11 & 43.93\% &18.28\% \\
		\hline
		PWCNet-fog~\cite{PWC} & 1.67 & 34.08\% &9.04\% \\
		\hline                    
		FlowNet2~\cite{FlowNet2} &5.92 &52.42\% &30.78\% \\
		\hline
		FlowNet2-defog~\cite{FlowNet2} &5.43 & 50.05\% &28.80\% \\
		\hline
		FlowNet2-fog~\cite{FlowNet2} &9.64 & 73.02\% &48.79\% \\
		\hline
		\textbf{Ours} &\textbf{1.60} & \textbf{28.31\%} &\textbf{8.45\%} \\
		\hline
		\hline
	\end{tabularx}
	\vspace{-0.15in}  
\end{table}

\vspace{0.3cm}
\noindent{\bf Quantitative Evaluation}
Since we target real dense fog images in our method and design, we do the evaluation on real images. 
EPE and ``bad pixel'' are commonly used metrics to measure the quality of optical flow. The definition of ``bad pixel'' follows to that of the KITTI dataset~\cite{menze2015CVPR}. Since the flow ground-truths in our evaluation are manually annotated, they might be inaccurate. To account for this, following the KITTI dataset~\cite{menze2015CVPR}, we compute ``bad pixel'' with its threshold parameter $\delta=\{3,5\}$, to allow for an inaccuracy of 3-5 pixels.
Table~\ref{table_chengdu_foggy} shows the evaluation result on our manually annotated real fog images. Our method has the best performance in terms of both EPE value and ``bad pixel'' numbers.
Table~\ref{table_vkitti_foggy} shows the results on synthetic fog images from the vKITTI dataset.  Since for synthetic fog images, we have accurate dense flow ground-truths, we compute ``bad pixel'' with $\delta=\{1,3\}$. While this is not our target, the evaluation shows that our method has comparable performance as the naive solutions.    


\begin{figure*}[t!]
	\begin{center}
		\captionsetup[subfigure]{labelformat=empty}
		{\includegraphics[width=0.25\textwidth, height= 1.8cm]{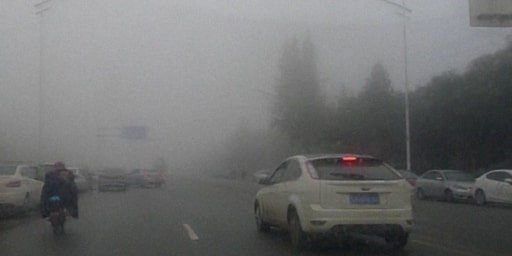}}\hfill
		{\includegraphics[width=0.25\textwidth, height= 1.8cm]{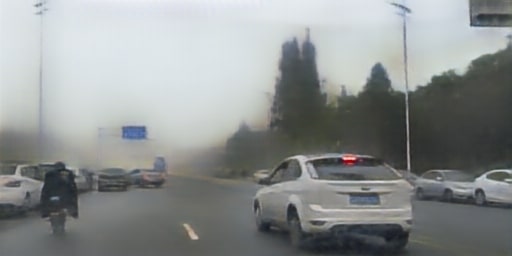}}\hfill
		{\includegraphics[width=0.25\textwidth, height= 1.8cm]{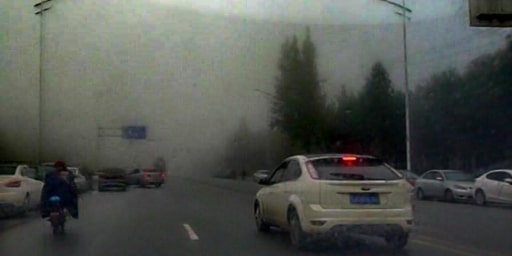}}\hfill
		{\includegraphics[width=0.25\textwidth, height= 1.8cm]{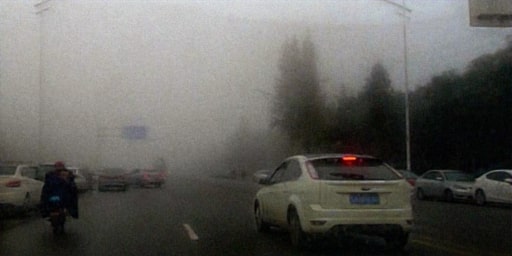}}\hfill
		
		\vspace{-0.14in}
		\subfloat[Input Image]{\includegraphics[width=0.25\textwidth, height= 1.8cm]{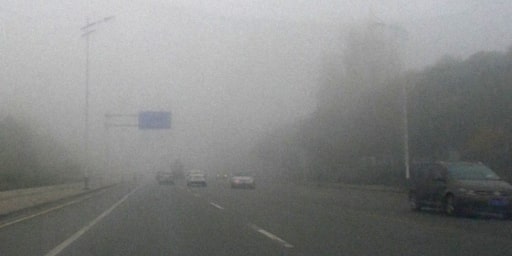}}\hfill
		\subfloat[\textbf{Our Result}]{\includegraphics[width=0.25\textwidth, height= 1.8cm]{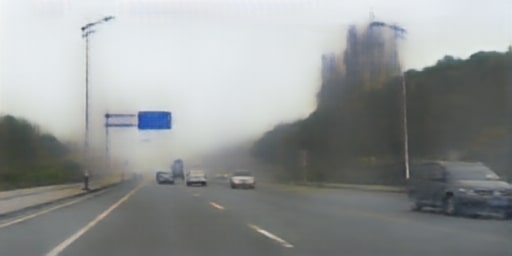}}\hfill
		\subfloat[Berman et al.~\cite{8540862}]{\includegraphics[width=0.25\textwidth, height= 1.8cm]{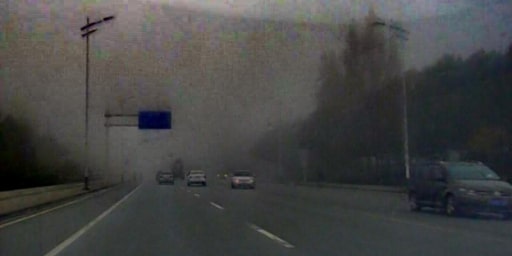}}\hfill
		\subfloat[EPDN~\cite{Qu_2019_CVPR}]{\includegraphics[width=0.25\textwidth, height= 1.8cm]{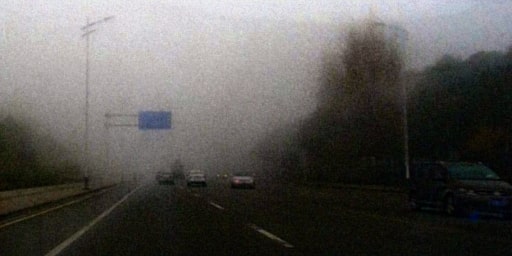}}\hfill\\	 	
		\vspace{-0.4cm}	
	\end{center}
	\caption{Qualitative defogging results on real foggy images. Although defogging is not our main target, we can observe that our method generates less artifacts than the state of the art methods do.}\label{qual_defog}
	\vspace{-0.15in}
\end{figure*} 

\begin{figure}[t!]
	\vspace{-0.125in}
	\begin{center}
		\captionsetup[subfigure]{labelformat=empty}	
		\subfloat{\includegraphics[width=0.155\textwidth]{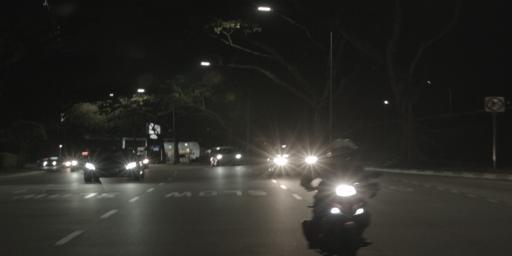}}\hfill
		\subfloat{\includegraphics[width=0.155\textwidth]{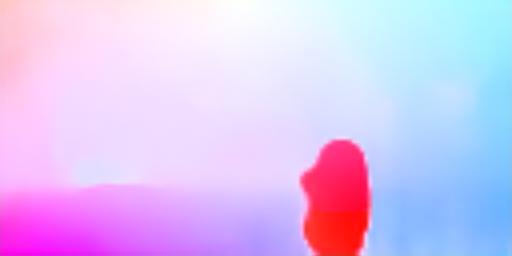}}\hfill
		\subfloat{\includegraphics[width=0.155\textwidth]{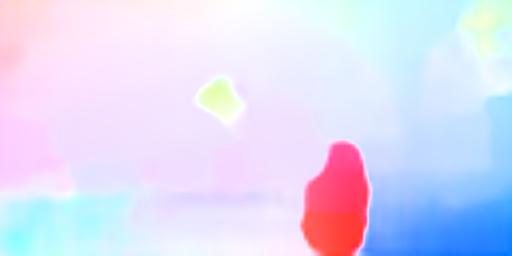}}\hfill\\
		\vspace{-0.12in}
		\subfloat[Input Image]{\includegraphics[width=0.155\textwidth]{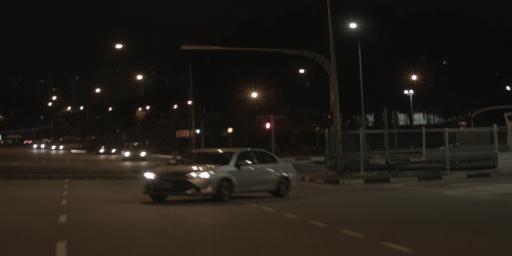}}\hfill
		\subfloat[\textbf{Our Result}]{\includegraphics[width=0.155\textwidth]{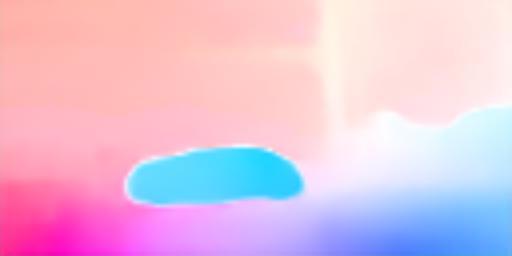}}\hfill
		\subfloat[PWCNet~\cite{PWC}]{\includegraphics[width=0.155\textwidth]{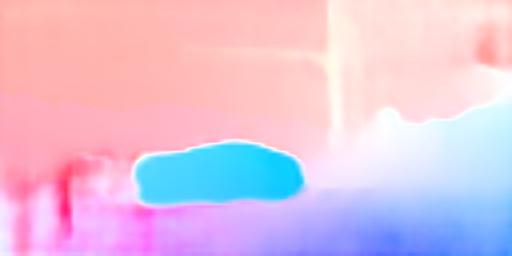}}\hfill\\	 	
		\vspace{-0.4cm}	
	\end{center}
	\caption{Qualitative methods on real nighttime images. As the results show, our method is not limited to fog, but can also work robustly for a different domain, such as nighttime.}\label{qual_night}
\end{figure} 

\begin{figure}[t!]
	\begin{center}
		\captionsetup[subfigure]{labelformat=empty}
		{\includegraphics[width=0.3333\columnwidth]{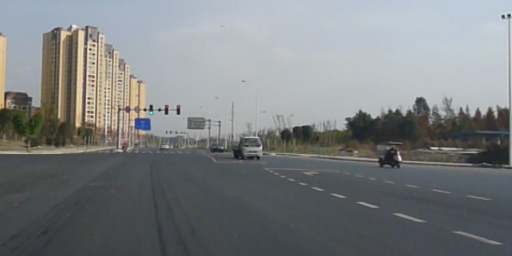}}\hfill
		{\includegraphics[width=0.3333\columnwidth]{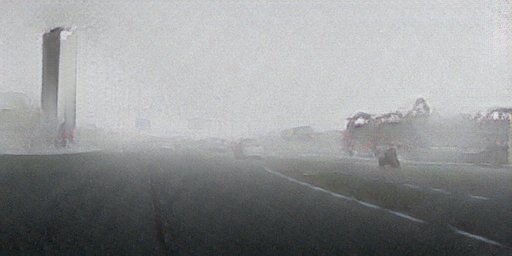}}\hfill
		{\includegraphics[width=0.3333\columnwidth]{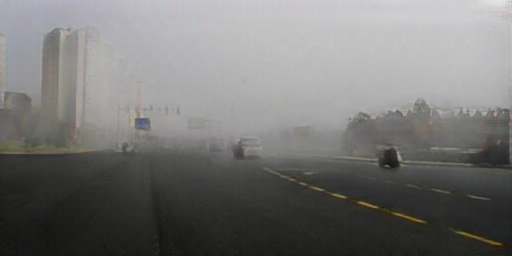}}\hfill\\	        \vspace{-0.14in}
		\subfloat{\includegraphics[width=0.3333\columnwidth]{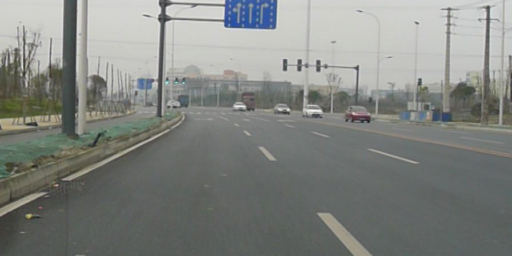}}\hfill
		\subfloat{\includegraphics[width=0.3333\columnwidth]{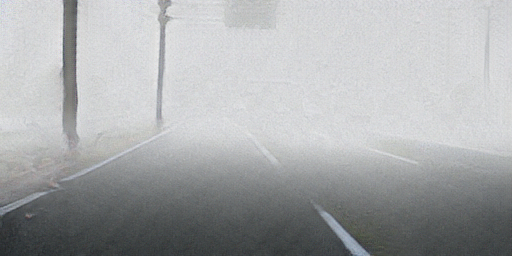}}\hfill
		\subfloat{\includegraphics[width=0.3333\columnwidth]{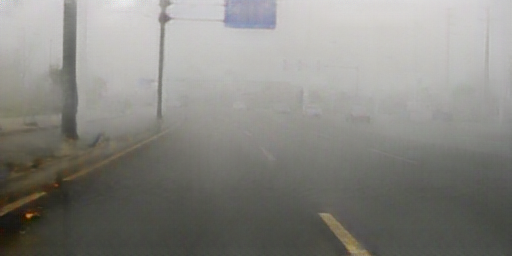}}\hfill
		\vspace{-0.4cm}
	\end{center}
	\caption{In each row, images from left-to-right show the input clean image, and the corresponding rendered fog images with and without the hazeline loss. The hazeline loss constrains our rendered fog images to avoid having fake colors.}\label{hazeline}
\end{figure}

\begin{figure}
	\begin{center}
		\captionsetup[subfigure]{labelformat=empty}
		{\includegraphics[width=0.3333\columnwidth]{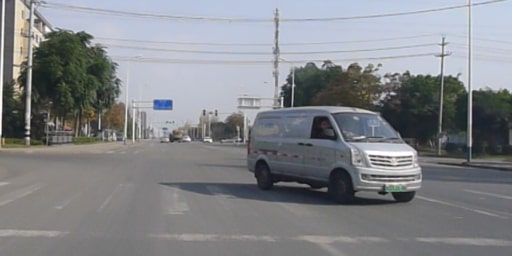}}\hfill
		{\includegraphics[width=0.3333\columnwidth]{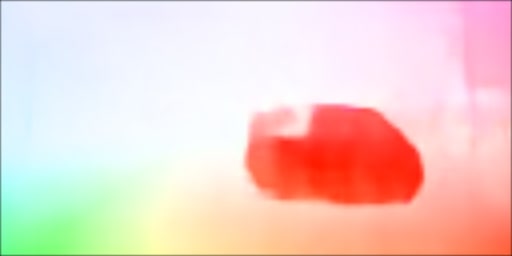}}\hfill 
		{\includegraphics[width=0.3333\columnwidth]{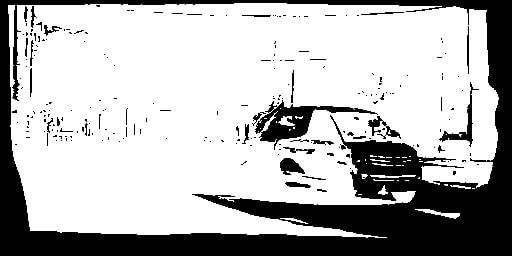}}\hfill\\
		\vspace{-0.14in}
		\subfloat[Input Clean Image]{\includegraphics[width=0.3333\columnwidth]{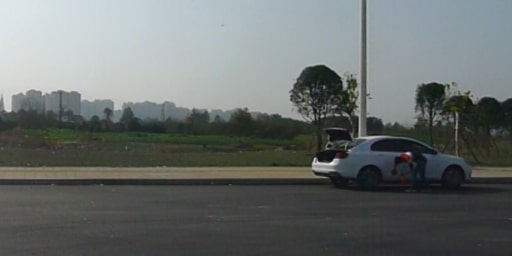}}\hfill
		\subfloat[Estiamted Flow]{\includegraphics[width=0.3333\columnwidth]{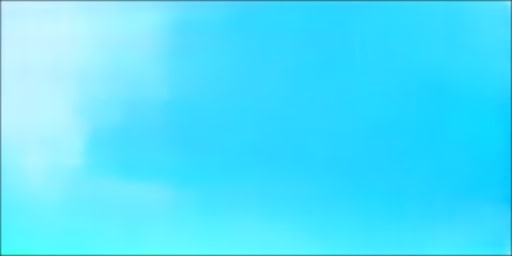}}\hfill
		\subfloat[Consistency map]{\includegraphics[width=0.3333\columnwidth]{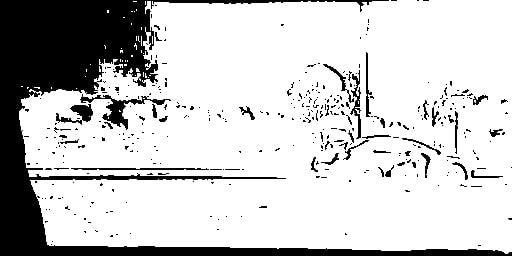}}\hfill
		\vspace{-0.4cm}
	\end{center}
	\caption{The photometric consistency masks correctly indicate the wrong flow estimations.}\label{consis_map}
	\vspace{-0.15in}
\end{figure}  

\vspace{0.3cm}
\noindent{\bf Qualitative Evaluation}
Fig.~\ref{qual} shows the qualitative results on real fog images. The first column shows the first image of the input image pair. All ground-truths in the second column are labeled manually by selecting rigid objects. The third column shows our results, and the other columns show the results of the baseline methods. As can be seen, our method in general performs better than the baseline methods, confirming our quantitative evaluation on the same data. Fig.~\ref{qual_defog} shows the qualitative defogging results on real foggy images. We compare our method with the state of the art of non-learning method Berman et al.~\cite{8540862} and learning-based method EPDN~\cite{Qu_2019_CVPR}. Although defogging is not our main target, we can observe that our method generates less artifacts.

While our method is designed for flow estimation under dense fog, we show that our method can also be applied to other domains, such as nighttime. For this, we use our entire training procedure as is, except for the hazeline loss described in Eq.~(\ref{eqn_HL_c_r}). The results are shown in Fig.~\ref{qual_night}. 

\section{Ablation Study}
Fig.~\ref{hazeline} shows the efficacy of our hazeline loss. It can constrain the color shifting between the clean images and the rendered fog images. We can observe that better fog images are generated by using the constraint.
Table~\ref{table_chengdu_foggy} shows the performance with and without the hazeline loss. Without the hazeline loss, our performance drops by 0.5 for EPE and 2-6\% on ``bad pixel'' rate. 

Fig.~\ref{consis_map} shows the binary photometric consistency masks. In the first row, our estimated optical flow has error on the minibus back window, and the mask can clearly show that area is inconsistent (black indicates inconsistent predictions, and white indicates consistent predictions). 
In the second row, the scene is static and the camera is moving. The optical flow is only generated by ego-motion. The estimated optical flow observably has errors on the left top corner. Our consistency mask also indicates the same.
The consistency mask and setting proper hyper-parameters (Sec.~\ref{sec:implementation}) are important for training stabilization. In our experiments, we find that the training loss can fail to converge if the consistency mask is not used.
We also check the efficacy of the domain transformation module. We observe that without this module (i.e. using only $E_f$ and $D_{of}$ in our network in Fig.~\ref{fig:network}), the performance of our method drops by 1.99 for EPE on real fog images.

\section{Conclusion}    

In this paper, we have proposed a semi-supervised learning method to estimate optical flow from dense fog images. 
We design a multi-task network that combines domain transformation and optical flow estimation. Our network learns from both synthetic and real data. The synthetic data is used to train our network in a supervised manner, and the real data is used in an unsupervised manner.
%
%
Our experimental results show the effectiveness of our method, which outperforms the state-of-the-art methods.

{\small
\bibliographystyle{ieee_fullname}
\bibliography{egbib}
}

\end{document}